# Dyna: A Method of Momentum for Stochastic Optimization

Zhidong Han

May 12, 2018

Livermore Software Technology Corporation (LSTC)
7374 Las Positas Road, Livermore, CA 94551
email: zdhan@lstc.com

**Abstract**

An algorithm is presented for momentum gradient descent optimization based on the first-order differential equation of the Newtonian dynamics. The fictitious mass is introduced to the dynamics of momentum for regularizing the adaptive stepsize of each individual parameter. The dynamic relaxation is adapted for stochastic optimization of nonlinear objective functions through an explicit time integration with varying damping ratio. The adaptive stepsize is optimized for each individual neural network layer based on the number of inputs. The adaptive stepsize for every parameter over the entire neural network is uniformly optimized with one upper bound, independent of sparsity, for better overall convergence rate. The numerical implementation of the algorithm is similar to the Adam Optimizer, possessing computational efficiency, similar memory requirements, etc. There are three hyper-parameters in the algorithm with clear physical interpretation. Preliminary trials show promise in performance and convergence.

## 1 Introduction

Gradient descent-based Backpropagation is one of the most widely used optimization techniques for training neural networks. In order to improve the convergence of the optimization methods, an adaptive stepsize is widely used to stabilize and speed up the learning process. The Adam Optimizer is one of the most popular methods for training neural networks.

A momentum-based method, *Dyna*, is presented in this paper with an accelerated gradient. This method is adapted from the explicit finite element method for solving the Newtonian Dynamics. A fictitious mass is introduced for each learning parameter to achieve a regularized angular frequency of each parameter, instead of unit mass. The dynamic relaxation with proper damping is used to perform the explicit time integration for faster convergence and reduced oscillation. The critical damping ratio of 1.0 can be used for better stability in the stochastic gradient case. This method can be reviewed as the combination of the Adam Optimizer [Kingma and Ba (2014)] and the Nesterov Momentum [Nesterov (1983)]. Its implementation is as straightforward as the Adam Optimizer, and determines adaptive learning rates for each individual parameter and each individual neural network layer. The pseudo-code is listed in Algorithm 1 for updating parameters. A simplified version is provided in Section 4.3 for implementing a rapid prototype based on the Adam Optimizer.

The theoretical background of the Newtonian Dynamics is briefed in Section 2 and its stochastic approach is formulated in Section 3. Section 4 provides some discussions on the numerical implementation. Some preliminary results are shown in Section 5 with some remarks in Section 6 for further investigation.

---

Algorithm 1: *Dyna*, a momentum optimizer for machine learning

---

**Neural network model:**
  $e(\theta)$ : error function
  $n^{\{l\}}$ : number of inputs of a neural network layer $\{l\}$
**Hyper parameters:**
  $\gamma$ : stepsize factor (defaulted $\gamma = 1$)
  $\beta \in [0, 1)$ : smoothing factor, as decay rate (defaulted $\beta = 0.9$ )
  $\zeta \in (0, 2]$ : damping ratio (defaulted $\zeta = 1$)
  $\hat{\omega}_\epsilon$ : minimum cut-off angular frequency (defaulted $\hat{\omega}_\epsilon = 10^{-8}$)
**Initial values:**
  $\theta_0 \leftarrow 0$ : initial parameter vector
  $\eta_0 \leftarrow 0$ : initial squared angular frequency vector
  $v_0 \leftarrow 0$ : initial velocity vector
  $t \leftarrow 0$ : initial timestep
  $\mu_0 \leftarrow 0$ : initial total weight
**While** $\theta$ not converged **do**
  $t \leftarrow t + 1$
  **for** each neuron network layer $\{l\}$ **do**
    $g_t \leftarrow \nabla_\theta \, e_t(\theta_t)$    gradients
    $\eta_t \leftarrow \beta \cdot \eta_{t-1} + (1 - \beta) \cdot |g_t|$    squared angular frequency estimate
    $\mu_t \leftarrow \beta \cdot \mu_{t-1} + (1 - \beta)$    total weight
    $\hat{\omega}_t \leftarrow \sqrt{\eta_t / \mu_t} + \hat{\omega}_\epsilon$    normalized angular frequency estimate
    $v_t \leftarrow \beta \cdot v_{t-1} - (1 - \beta)/(2\zeta) \cdot g_t / \hat{\omega}_t$    velocity estimate
    $\hat{v}_t \leftarrow v_t / \mu^t$    normalized velocity estimate
    $\alpha_t \leftarrow 2\gamma / n^{\{l\}}$    learning rate
    $\theta_t \leftarrow \theta_{t-1} + \alpha_t \cdot \hat{v}_t / \hat{\omega}_t$    parameters
  **end for**
**Return** $\theta$

---

The damping ratio $\zeta$ may be set to 0.5 or less (as under damped) for a fast convergence rate at the beginning. It may be increased to 1.0 (as critically damped) or higher (as over damped) during learning process for better convergence.

---

## 2  Momentum Algorithms

The momentum algorithms presented in this section have been well studied and formulated over the course of several decades within the computational mechanics society. They have been widely adapted for the explicit finite element analysis (FEA). As one of the most popular commercial codes, LS-DYNA© originated from the 3D FEA program DYNA3D, developed by Dr. John O. Hallquist at Lawrence Livermore National Laboratory (LLNL) in 1976. The code and its application in car crash simulation have been presented in the Computer History Museum located in Mountain View, CA, USA, at  *http://www.computerhistory.org/makesoftware/exhibit/car-crash-simulation*

   The fundamental concept of the momentum algorithms is illustrated with the use of a single coordinate system. The conclusions drawn here are based on the enormous research work which is not listed in the present paper. For further theoretical details and a comprehensive list of references refer to the LS-DYNA Theory Manual [Hallquist (2018)] which is available online at *http://www.lstc.com/download/manuals*.

## 2.1 A Spring Mass System

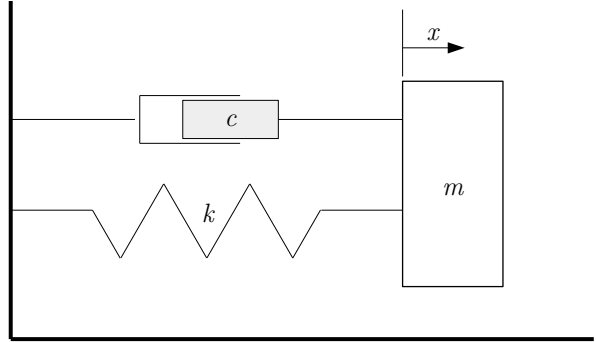

Figure 1: a spring mass system

Consider a mass $m$ at the end of a spring as in Fig. 1. The spring exerts a force on the mass if the spring is stretched from its natural length, as

$$F = ma = m\frac{dv}{dt} = m\frac{d^2x}{dt^2} = -kx \qquad (1)$$

where $k$ is the spring stiffness (as the gradient), $a$ is the acceleration, $v$ is the velocity, and $x$ is the coordinate.

By solving this second-order linear differential equation, the general solution of the periodic motion can be expressed as a simple harmonic motion,

$$x(t) = A\cos(\omega t + \phi) \qquad (2)$$

where $\omega = \sqrt{k/m}$ is the angular frequency.

A damping force can be introduced if the motion of a mass is subject to a frictional force. We assume that the damping force is proportional to the velocity of the mass and acts in the direction opposite to the motion. Eq. 1 can be revised as,

$$ma + cv + kx = 0 \qquad (3)$$

where $c$ is the damping coefficient. By defining the damping ratio $\zeta = c/2\sqrt{mk} = c/2m\omega$, we have the second-order linear differential equation for the damped motion, as

$$a + 2\zeta\omega v + \omega^2 x = 0 \qquad (4)$$

As shown in Fig. 2, the mass returns to its equilibrium position very fast without oscillating if the system is critically damped when ($\zeta = 1$).

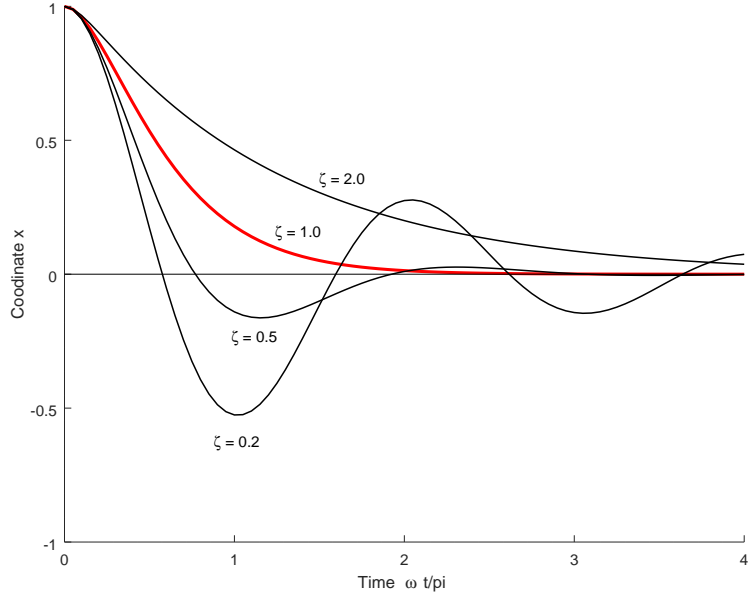

Figure 2: The effect of damping ratio

## 2.2 Explicit Time Integration

Considering explicit time methods, the central difference method is still a widely used scheme. It has the largest time step stability limit of any second-order accurate explicit method.

Without losing the generality, we take a constant time step, $h$, at any point in time, $t$. The central difference time integration updates the coordinate to time $t + h$ as

$$\begin{aligned} a_t &= -k/m \cdot x_t \\ v_{t+h/2} &= v_{t-h/2} + a_t \cdot h \\ x_{t+h} &= x_t + v_{t+h/2} \cdot h \end{aligned} \quad (5)$$

The accuracy and stability of the central difference time integration depend upon the ratio of the time step, $h$, to the highest angular frequency of the system with multiple coordinates, $\omega_{max}$, as

$$h \leq h_{critical} \equiv 2/\omega_{max} \quad (6)$$

For a spring mass system, the critical stepsize $h_{critical} = 2/\omega$. It increases to $2/(\omega\sqrt{1-\zeta^2})$ if the system is damped and the explicit time integration becomes more stable.

## 2.3 Dynamic Relaxation

The dynamic relaxation algorithm includes the damping in Eq. 3 for the central difference time integration. The damping coefficient is selected to increase the convergence speed towards the equilibrium position without oscillating.

With the central difference scheme, we have

$$\begin{aligned} a_t &= (v_{t+h/2} - v_{t-h/2})/h \\ v_{t+h/2} &= (x_{t+h} - x_t)/h \end{aligned} \quad (7)$$

and $v_t$ by averging velocity over $[t - h/2, t + h/2]$,

$$v_t = (v_{t+h/2} + v_{t-h/2})/2 \tag{8}$$

for updating velocity $v_{t+h/2}$ with Eq. 4, as

$$v_{t+h/2} = \beta \cdot v_{t-h/2} - (1 - \beta)/(2\zeta) \cdot \omega \cdot x \tag{9}$$

where by definition,

$$\beta = 1 - 2\zeta\omega h/(1 + \zeta\omega h) \tag{10}$$

The central difference scheme is unconditionally stable without oscillating if the timestep is smaller than the critical timestep defined in Eq. 6 with a corresponding damping ratio $\zeta = 1$.

## 3 Stochastic Optimization

The momentum algorithms in Section 2 work for physical models in which the mass, stiffness and damping coefficient are defined as known parameters. In the stochastic gradient case, most model parameters are unknown. We need angular frequencies to determine the critical damping coefficients and time step for stable convergence. In this section, we recover the model parameters from Backpropagation.

### 3.1 Adaptive Fictitious Mass and Angular Frequency Estimate

The Backpropagation algorithm applies the chain rule and transmits intermediate errors, $e(\theta)$, backwards through the neural network layers. The gradients are computed for the current layer, but not transmitted backwards. It implies that we only have errors serving as the force $kx$ in the momentum method in Eq. 1, instead of the stiffness $k$.

Let $e(\theta)$ be a stochastic scalar function, and the expectation of its absolute value associated with parameter $\theta$ as

$$\eta = \mathbb{E}[|e(\theta)|] \tag{11}$$

and its square root as the fictitious angular frequency estimate,

$$\hat{\omega} \equiv \sqrt{\mathbb{E}[|e(\theta)|]} \tag{12}$$

for updating velocity in Eq. 9 with $\omega \cdot x = e(\theta)/\hat{\omega}$. Essentially, we apply a fictitious mass estimate of $\mathbb{E}[|e(\theta)|]/k$ to parameter $\theta$ in which the stiffness $k$ remains unknown. A larger $e(\theta)$ results to a larger fictitious mass. This is a desirable property for regularizing the angular frequencies for all parameters in a neural network model. It needs to be pointed out that the true undamped angular frequency is $k/\hat{\omega}$ which is expected to be proportional to $\sqrt{k/x}$. It needs to be reduced by applying the critical damping besides normalizing.

### 3.2 Velocity Estimate and Normalization

In the stochastic gradient case, hyper-parameter $\beta$ in Eq. 10 is used for updating the exponential moving average of the squared angular frequency. It is the recommended to set $\beta = 0.9$ as $\beta_1$ in the Adam Optimizer. We have the estimate as

$$\eta_t = \beta \cdot \eta_{t-1} + (1 - \beta) \cdot |e(\theta)| \tag{13}$$

which can be regarded as a weight average over the previous values. It needs to be normalized to reflect its correct value. The total weight can be accumulated as

$$\mu_t = \beta \cdot \mu_{t-1} + (1 - \beta) \tag{14}$$

and the normalized angular frequency estimate is obtained,

$$\hat{\omega}_t = \sqrt{\eta_t/\mu_t} + \hat{\omega}_\epsilon \tag{15}$$

where $\hat{\omega}_\epsilon = 10^{-8}$ is the low cut-off value of the fictitious angular frequency estimate.

With Eq. 9, the velocity estimate can be updated with exponential moving average and a damping ratio, as,

$$v_t = \beta \cdot v_{t-1} + (1 - \beta)/(2\zeta) \cdot e(\theta)/\hat{\omega}_t \tag{16}$$

and normalized by using the same weight in Eq. 14, as

$$\hat{v}_t = v_t/\mu_t \tag{17}$$

Instead of remaining constant, the damping ratio may adaptively vary between under-damping with $(\zeta < 1)$ for a faster convergence rate, and over-damping with $(\zeta > 1)$ for stabler convergence.

## 3.3 Adaptive Timestep

With the normalized angular frequency estimate in Eq. 15, it is straightforward to set the adaptive timestep $h = 2/\hat{\omega}_t$. Let $n$ be the number of inputs in a neural network layer. We may update all parameters evenly within a layer by rescaling $h$ to $h/n$. Therefore, each parameter can be updated with Eq. 5, with the use of the normalized velocity estimate in 17, as

$$\theta_t = \theta_{t-1} + \alpha_t \cdot \hat{v}_t/\hat{\omega}_t \tag{18}$$

where by definition, $\alpha_t = 2\gamma/n$ is the learning rate with an optional stepsize factor $\gamma$ that defaults to 1.0.

# 4 Numerical Implementation

The algorithm is formulated for the stochastic optimization based on the Newtonian Dynamics. The physical interpretation of the hyper parameters in the algorithm can be used to build some numerical strategies for machine learning.

## 4.1 Total Weight and Restart

Let hyper-parameter $\beta$ be constant. The total weight $\mu_t$ in Eq. 14 at timestep $t$ can be obtained as,

$$\mu_t = (1-\beta) \sum_{i=1}^{t} \beta^{t-1} = (1 - \beta^t) \tag{19}$$

which is the correction factor of the initialization bias in the Adam Optimizer. It implies that its exponential moving average is under damped with a damping ratio $\zeta = 0.5$. Therefore, it is reasonable to let $\zeta$ start from 0.5, as the Adam Optimizer is fairly stable.

With Eqs. 13, 14 and 16, the squared angular frequency estimate and velocity share the same synchronous total weight. It is possible to zero or rescale the existing estimates as well the total weight within a neural network layer to restart or warm restart the learning process. This operation can be performed within the same time step whenever the objective function increases.

## 4.2 Effective Adaptive Timestep

With Eqs. 13 and 16, the effective timestep is bounded by $\alpha_t$ as $|\hat{v}_t| \leq \hat{\omega}_t$ for less sparse cases with $\zeta \geq 0.5$. This is most likely true at the beginning. When parameters are close to the global optimum, most gradients are expected to be zero or very small. We may need to introduce some noise to increase the momentum. For example, the random dropout can be applied once the objective function decreases slowly. Assuming the angular frequency estimate $\eta_{t-1} = 0$, and the velocity estimate $v_{t-1} = 0$, the update process in timestep $t$ is carried out as,

$$
\begin{aligned}
\mu_t &= \beta \cdot \mu_{t-1} + (1 - \beta) \\
\eta_t &= (1 - \beta) \cdot |e(\theta)| \\
\hat{\omega}_t &= \sqrt{(1 - \beta) \cdot |e(\theta)|} / \sqrt{\mu_t} \\
|v_t| &= (\sqrt{(1 - \beta) \cdot |e(\theta)|} \cdot \sqrt{\mu_t}) / (2\zeta) \\
|\hat{v}_t| &= (\sqrt{(1 - \beta) \cdot |e(\theta)|} / \sqrt{\mu_t}) / (2\zeta) = \hat{\omega}_t / (2\zeta)
\end{aligned}
\quad (20)
$$

which shows that the effective timestep is still bounded by $\alpha_t$ with $\zeta \geq 0.5$. This is a desirable property to keep the algorithm stable under sudden impact in both sparse or less sparse cases. However, less damping with $\zeta < 0.5$ may help in increasing the momentum for speeding up the learning process if necessary. So the damping ratio really plays an important role in tuning the overall learning process.

The third hyper parameter $\hat{\omega}_\epsilon$ is the low cut-off value of the angular frequency. When the maximum angular frequency estimate, $\max(\hat{\omega}_t)$, in a layer becomes extremely small, $\hat{\omega}_\epsilon$ may be reduced to a small number of several orders lower for this layer, to maintain the effective timestep. Otherwise, the effective timestep decreases because $\hat{\omega}_t$ is much larger than $|\hat{v}_t|$. In general, $\hat{\omega}_\epsilon = 10^{-8}$ is small enough for most cases.

## 4.3 Prototype Implementation

The Adam Optimizer has been implemented in machine learning packages and configured for various applications. A prototype version of the present algorithm can be implemented by replacing the last five lines in the loop in Algorithm 1 in the Adam Optimizer with the following lines:

$$
\begin{aligned}
v_t &\leftarrow \beta_1 \cdot v_{t-1} + (1 - \beta_1) \cdot |g_t| \\
\hat{v}_t &\leftarrow \sqrt{v_t/(1 - \beta_1^t)} + \epsilon \\
m_t &\leftarrow \beta_1 \cdot v_{t-1} + [(1 - \beta_1)/(2\beta_2)] \cdot g_t / \hat{v}_t \\
\hat{m}_t &\leftarrow m_t/(1 - \beta_1^t) \\
\theta_t &\leftarrow \theta_{t-1} - \alpha \cdot \hat{m}_t / \hat{v}_t
\end{aligned}
\quad (21)
$$

in which $\beta_2$ is used as the damping ratio that defaults $\zeta = 0.999$.

# 5 Experimental Results

Logistic regression was studied using the MNIST dataset by running a Python code by Bondarenko (2017), to classify the class label directly on the 784 dimension image vectors. Lines in function *logreg* (in file *main.py*) for extracting feature vectors are disabled. The prototype implementation in Section 4.3 is coded in function *adam::update* (in file *ml_mnist\optimizer.py*). The hyper parameters are set to $n\_batches = 96$, $L2 = 0.000016$, $\beta_1 = 0.9$, and $learning\_rate = 0.00255 = 2/784$ with no decay.

The first two tests were carried out with constant damping ratios of $\zeta = 1.0$ (critically damped) and $\zeta = 0.5$ (under-damped), respectively. The third test was carried out with a varying damping ratio increasing from 0.5 to 1.0 in 10 epochs, which is coded as

$$1/(2\zeta) = 1.0 - 0.5\min(1, (t/T)^2) \qquad (22)$$

where $T = 10 \times n\_batches$.

The Adam Optimizer was tested with a constant learning rate of 0.001 as a reference. In Fig. 3, the present algorithm, the Dyna Optimizer, shows the similar convergence as the Adam Optimizer. Among three tests of the present algorithm, the vary damping ratio gives better performance as expected. The accuracies of all tests after 50 epochs are listed in Table 1, and the learning curves are shown in Fig. 4.

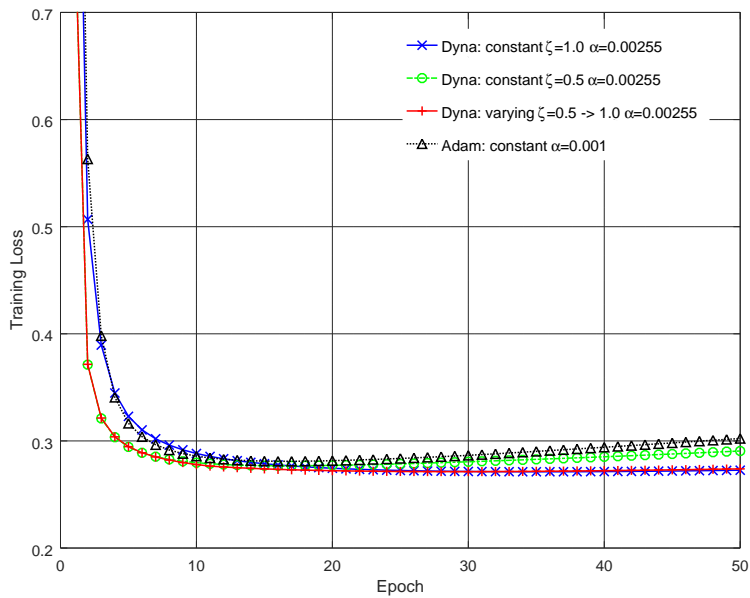

Figure 3: The effect of damping ratio on training loss

Table 1: Accuracies of logistic regression on MNIST images

| Algorithm | tr. acc. | val. acc. | test acc. | hyper parameters |
|---|---|---|---|---|
| Dyna | 0.9428 | 0.9368 | 0.9304 | constant $\zeta = 1.0$, l.r.=0.00255 |
| Dyna | 0.9448 | 0.9349 | 0.9306 | constant $\zeta = 0.5$, l.r.=0.00255 |
| Dyna | 0.9437 | 0.9349 | 0.9307 | varying $\zeta = 0.5 \to 1.0$, l.r.=0.00255 |
| Adam | 0.9435 | 0.9340 | 0.9286 | constant l.r.=0.001 |

*Note: all trainings were terminated after 50 epochs.*

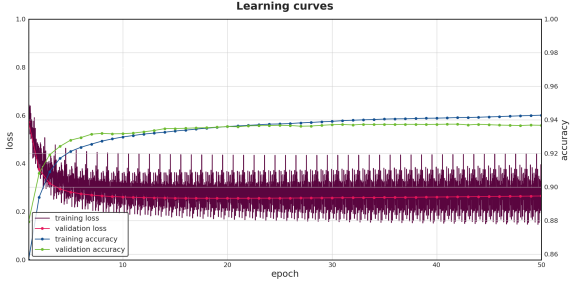
(a) Dyna: constant $\zeta = 1.0$, l.r.=0.00255

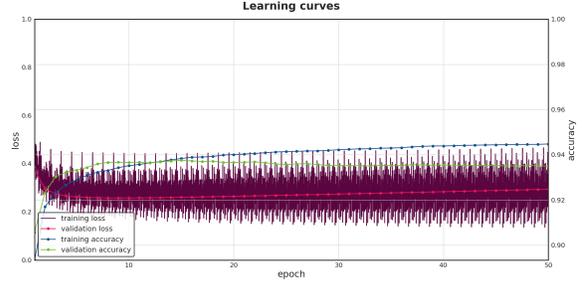
(b) Dyna: constant $\zeta = 0.5$, l.r.=0.00255

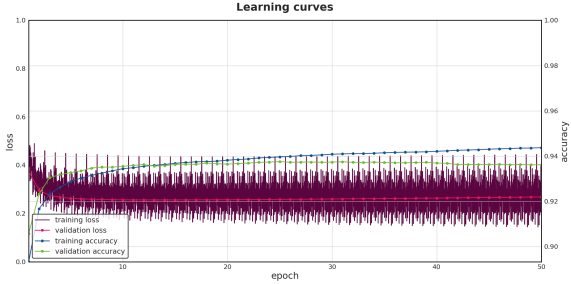
(c) Dyna: varying $\zeta = 0.5 \to 1.0$, l.r.=0.00255

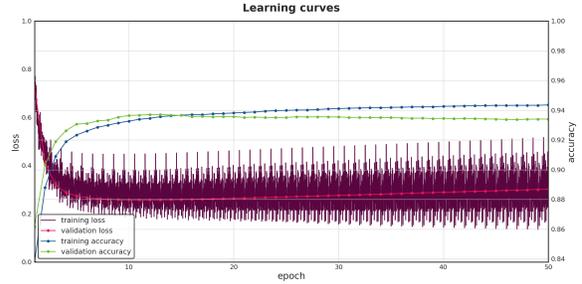
(d) Adam: constant l.r.=0.001

Figure 4: The effect of damping ratio on learning curve

## 6 Some Remarks

There are several million lines of Fortran/C/C++ code in LS-DYNA[©]. It needs to run over thousands of CPUs for an explicit analysis of a state-of-art car model with 100 million elements. It becomes difficult to profile over 50 thousand functions/subroutines in the code besides third party libraries. With the AI technology in mind, the author started to read *Deep Learning* by Goodfellow et al. (2016) with great interest, especially when the Newtonian Dynamics was introduced in Section 8.3. However, the authors of *Deep Learning* made a comment that *"Unfortunately, in the stochastic gradient case, Nesterov momentum does not improve the rate of convergence."* Thanks to Prof. Andew Ng at Standform, who explained the algorithms including the Adam Optimizer in a simple straightforward manner in his online courses *Machine Learning, Neural Networks and Deep Learning* and others on COURSERA[©] [Ng (2012, 2017)]. With the connections between the Newtonian momentum and LS-DYNA[©], the author have come to the realization that many algorithms in the explicit FEA could be adapted for machine learning, especially numerical techniques pioneered by Dr. John O. Hallquist making the code computationally efficient and robust.

The algorithm in the present paper, *Dyna*, is still under investigation, although the results of preliminary trials are promising. However, it requires modern neural network models and other resources, which are not available to the author, for further investigation. Therefore, the author decided to publish the algorithm in its early stage.

The author would like to thank David Han at UC Berkeley for his thoughtful discussions and contributions to the paper.